\documentclass{article}
\usepackage{arxiv}
\usepackage{array}
\usepackage{amsmath}
\usepackage{xcolor}
\usepackage{listings}
\usepackage{tikz}
\usetikzlibrary{arrows.meta,positioning}

\newcommand{\N}{\mathsf{N}}
\newcommand{\T}{\mathsf{T}}
\newcommand{\F}{\mathsf{F}}
\newcommand{\B}{\mathsf{B}}

\newcommand{\code}[1]{\texttt{#1}}
\newcommand{\adm}{\mathop{\rm adm}}

\usepackage{amsthm}
\newtheoremstyle{agra}{3pt}{3pt}{\normalfont}{0pt}
  {\itshape\color{agrablue}}{:}{0.5em}
  {\thmname{#1}\thmnumber{ #2}\thmnote{ (#3)}}
\theoremstyle{agra}
\newtheorem{theorem}{Theorem}
\newtheorem{lemma}{Lemma}
\makeatletter
\renewenvironment{proof}[1][\proofname]{\par
  \pushQED{\qed}\normalfont\topsep6\p@\@plus6\p@\relax
  \trivlist\item[\hskip\labelsep\itshape #1:]\ignorespaces
}{\popQED\endtrivlist\@endpefalse}
\makeatother
\newenvironment{IEEEproof}[1][\proofname]{\begin{proof}[#1]}{\end{proof}}
\usepackage{enumitem}

\usepackage[most]{tcolorbox}
\newif\iftheorembox
\theoremboxtrue
\iftheorembox
  \tcolorboxenvironment{theorem}{enhanced, breakable, sharp corners,
    frame hidden, boxrule=0pt, colback=agralightblue!30,
    left=4pt, right=4pt, top=3pt, bottom=3pt, boxsep=0pt,
    before skip=4pt, after skip=4pt}
  \tcolorboxenvironment{lemma}{enhanced, breakable, sharp corners,
    frame hidden, boxrule=0pt, colback=agralightblue!30,
    left=4pt, right=4pt, top=3pt, bottom=3pt, boxsep=0pt,
    before skip=4pt, after skip=4pt}
\else
  \AtBeginEnvironment{theorem}{\setlength{\parindent}{0pt}}
  \AtBeginEnvironment{lemma}{\setlength{\parindent}{0pt}}
\fi

\usepackage{soul}
\newif\ifreviewmarks
\reviewmarksfalse
\definecolor{revred}{HTML}{C00000}
\definecolor{revorange}{HTML}{D96C00}
\definecolor{revyellow}{HTML}{FFE14D}
\definecolor{revblue}{HTML}{1F5FBF}

\definecolor{agrablue}{HTML}{004D8E}
\definecolor{agralightblue}{HTML}{89B4E1}

\lstdefinestyle{agra}{
  language=C,
  basicstyle=\ttfamily\footnotesize,
  keywordstyle=\color{agrablue}\bfseries,
  commentstyle=\color{agrablue!70}\itshape,
  backgroundcolor=\color{agralightblue!30}, 
  frame=none, 
  xleftmargin=3pt, xrightmargin=3pt, framexleftmargin=3pt, framexrightmargin=3pt,
  columns=fullflexible, keepspaces=true, upquote=true,
  aboveskip=0pt, belowskip=0pt
}

\tikzset{
  agrabox/.style={draw=none, fill=agralightblue,
    minimum height=7mm, align=center, inner sep=2pt,
    font=\footnotesize},
  agraarrow/.style={-{Stealth[length=2mm]}, draw=agrablue, line width=0.6pt},
  agralabel/.style={font=\scriptsize, text=agrablue, inner sep=1pt}
}

\title{Secure Formal Raising: Protecting the Development\protect\\of AI through Executable Properties}
\renewcommand{\shorttitle}{Secure Formal Raising}  
\date{}
\author{Rolf Drechsler\\
  University of Bremen and DFKI\\
  Bremen, Germany\\
  \texttt{drechsler@uni-bremen.de}
  \And
  Christina Plump\\
  University of Bremen and DFKI\\
  Bremen, Germany\\
  \texttt{cplump@uni-bremen.de}}

\begin{document}
\maketitle

\begin{abstract}
Model raising revisits Turing's developmental view of machine intelligence.
Formal Raising (FR) has recently been proposed to complement this view
with explicit behavioral properties.
This paper extends that proposal to the \emph{security} of the raising environment.
Secure Formal Raising (SFR) models policy integrity, authorization,
and confidential data flow using four-valued evidence that preserves unknown and
inconsistent assessments.
We prove six security invariants and show that they force admission
to accept only confirmed evidence.
A plain-C reference implementation satisfies the invariants
in a bounded exhaustive exploration,
which also exposes a confused-deputy gap for tool calls
that a provenance requirement closes.

\end{abstract}


\section{Introduction}\label{sec:01-introduction}
How should an intelligent machine acquire its behavior?
In 1950, Turing proposed educating a comparatively simple ``child machine''
instead of directly programming an adult-like intelligence~\cite{turing}.
He distinguished its initial construction from the education through which it
develops.
This developmental perspective makes interaction with teachers part of the
system's design, rather than merely a source of test inputs.

Aydin et~al.\ recently revived this perspective for contemporary AI through
\emph{model raising}~\cite{raising}. 
They propose shaping values with capabilities early on,
using first-person experience, social interaction, and a scaffolded curriculum.
Their proposal motivates bringing behavioral commitments into development instead of 
relying on later alignment.
It does not itself provide an executable specification of those commitments.

\emph{Formal Raising} (FR) is a newly proposed complement to this developmental
view~\cite{ethics}.
It represents selected requirements as explicit, machine-checkable properties
that participate in evaluation and action selection.
Hard rules exclude inadmissible alternatives, while declared priorities rank
competing preferences.
The preceding FR work primarily develops ethical examples;
it is a starting point for further investigation rather than an established
account of AI development.

This paper builds on that proposal by addressing the integrity of the raising
environment.
A teaching message may change a curriculum, a retrieved document may influence
an action, and a tool call may disclose information.
Who may change the policy, and which operations may a developing system execute?
Ethical evaluation alone does not establish teacher authenticity or
authorization.
The proposed \emph{Secure Formal Raising} (SFR) model therefore enforces
protection against unauthorized modification, execution,
and disclosure at explicit transition boundaries.

The environment is modeled as a state machine, making the approach accessible
through familiar verification concepts.
Four-valued evidence preserves confirmed, refuted, unknown,
and inconsistent assessments of identity and provenance.
Only unambiguous confirmation satisfies a hard requirement,
while unresolved evidence supports differentiated feedback.
The contribution combines this evidence layer with policy integrity,
limited authorization, preference ranking, and a self-contained plain-C
reference implementation.
Building on established access-control mechanisms,
we prove six security invariants for this model,
show that they require admission on confirmed evidence only,
and check the implementation against them by a bounded exhaustive exploration
of 6,262 reachable states and 72,329,329 transitions.
The exploration finds no violation,
but 21,708 admitted tool calls and exports with unconfirmed source evidence
reveal a confused-deputy gap, which a provenance requirement closes.

Section~\ref{sec:02-background} relates SFR to prior work, while 
Sec.~\ref{sec:03-sfr} defines the formal model to ensure security guarantees. 
Section~\ref{sec:04-implementation} presents the implementation
and checks it against the guarantees.
Section~\ref{sec:05-discussion} discusses limitations,
and Sec.~\ref{sec:06-conclusion} concludes.

\section{Background and Related Work}\label{sec:02-background}

SFR builds on four lines of work: FR, classical access control and
information flow,
four-valued logic in access control, and runtime enforcement for agents and
learners.

FR evaluates candidate actions against declared behavioral properties;
priorities resolve
competing preferences, and hard properties are non-defeasible~\cite{ethics}.
It continues
work on formally verifying the choices of autonomous systems~\cite{dennis}.
SFR applies
the same separation to the environment: teachers may change preferences,
but no learning
action may change the mechanism that admits it.

The security mechanisms of SFR are standard.
Admission follows the principles of fail-safe
defaults and complete mediation~\cite{saltzer}.
Roles restrict operations as in role-based
access control~\cite{sandhu}.
The flow check is the two-point instance of the lattice model
of information flow~\cite{denning}; it constrains explicit transfers only and
does not
establish noninterference~\cite{goguen}.
A program that acts with its own authority on
behalf of a less trusted party is a \emph{confused deputy}~\cite{hardy}.
We will show in Sec.~\ref{sec:03d-guarantees} that this situation is relevant in SFR and how
to prevent it.
Note that we do not consider availability and cryptographic mechanisms in this work.

Belnap's logic represents incomplete and inconsistent information by four
values~\cite{belnap}.
Ginsberg generalized its two orders, truth and knowledge,
to bilattices~\cite{ginsberg},
and Arieli and Avron base consequence relations on sets of designated
values~\cite{arieliavron}.
Bruns and Huth use Belnap's logic to compose access-control policies,
where $\B$ marks a conflict between
policies and $\N$ a gap~\cite{brunshuth}.
XACML's decisions \emph{NotApplicable} and
\emph{Indeterminate} play similar roles~\cite{xacml}.
SFR uses the same lattice for a
different purpose: the values describe \emph{evidence about a request},
namely whether
its actor and source are confirmed, rather than combined policy decisions.
The hard
properties themselves are fixed and are not composed.

Security automata enforce safety properties by monitoring
execution~\cite{schneider},
and runtime verification checks executions against formal
specifications~\cite{leucker}.
The closest analogue for learning systems is a \emph{shield},
which blocks unsafe actions
of a reinforcement learner~\cite{shield}.
SFR shields the protected transitions of the
environment, including updates to the policy that guides learning.
Specifications of this
kind can be expressed as assertions in property languages such as
PSL~\cite{psl}.

Indirect prompt injection lets retrieved content control LLM-integrated
applications~\cite{injection}.
AgentDojo provides a benchmark of such attacks
on tool-using agents~\cite{agentdojo}, and CaMeL separates control flow from
data flow and tracks provenance with capabilities~\cite{camel}.
SFR does not
track provenance itself; it assumes provenance evidence is supplied and
specifies how evidence of different quality gates actions.

\section{Secure Formal Raising}\label{sec:03-sfr}
This section defines SFR, which protects the preference policy:
the environment and its events (Sec.~\ref{sec:03a-model}),
the semantics of the model (Sec.~\ref{sec:03b-semantics}),
scenarios that illustrate it (Sec.~\ref{sec:03c-scenarios}),
and the guarantees it provides (Sec.~\ref{sec:03d-guarantees}).

\subsection{Environment Model}\label{sec:03a-model}
Actors have one of three roles: \emph{teacher}, \emph{student},
or \emph{external}.
Teachers propose policy updates and grant
limited authorization.
Students propose learning-related tool use or export.
External actors supply content and propose operations but cannot acquire an
internal role by stating so in the text.

Four operations keep the model small: \code{OBSERVE}
is an inert local action; \code{CHANGE\_POLICY} proposes a replacement
policy; \code{TOOL\_CALL} interacts with an external resource;
\code{EXPORT\_DATA} transfers data.
Tool calls and exports are
the \emph{protected} operations.
A candidate action consists of
its operation, actor, role, identity evidence, source evidence,
input and destination labels, expected policy version,
two preference costs, and, for updates, a replacement policy.
Evaluating a candidate never changes the state.

\emph{State and events.}
The state is
\begin{equation}\label{eq:state}
 s=(t,v,\ell,p_{\rm priv},p_{\rm prog},\pi),
\end{equation}
with logical clock $t$, policy version $v$,
permit lifetime $\ell$, preference priorities $p_{\rm priv}$ and $p_{\rm prog}$,
and permit record
\begin{equation}\label{eq:permit}
 \pi=(\mathit{active},\mathit{actor},\mathit{kind},v_\pi,d).
\end{equation}
The policy $(v,\ell,p_{\rm priv},p_{\rm prog})$ starts as $(1,2,40,30)$.
Priorities lie in $[0,100]$ and lifetimes in $[1,100]$.
The model holds a single permit; a new grant replaces it.

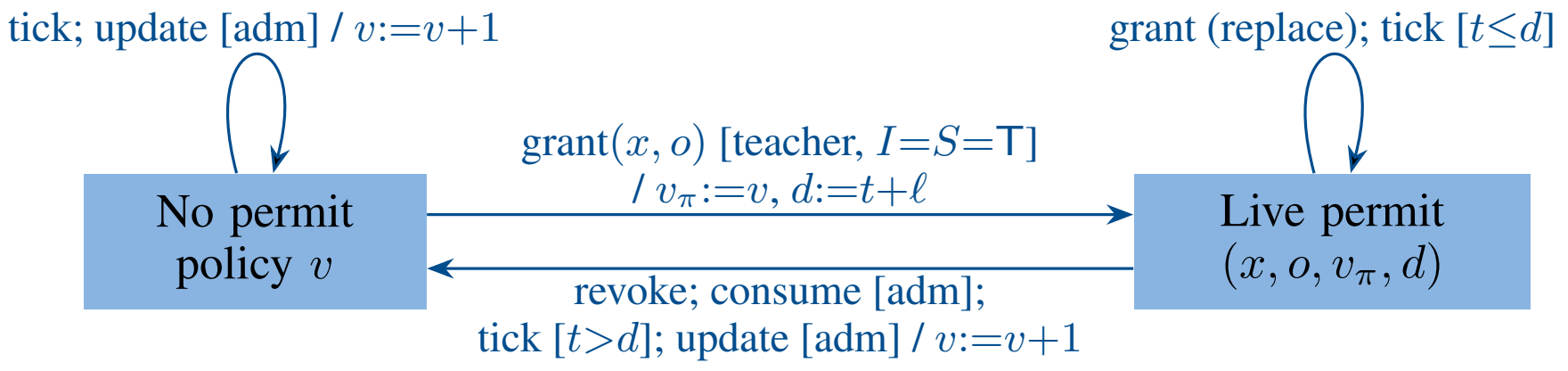
\begin{figure}[t]
\centering
\begin{tikzpicture}[agrabox/.append style={font=\normalsize, minimum height=10mm},
                    agralabel/.append style={font=\small}]
  \node[agrabox, text width=24mm] (none) at (0,0) {No permit\\policy $v$};
  \node[agrabox, text width=28mm] (live) at (80mm,0) {Live permit\\$(x,o,v_\pi,d)$};
  \draw[agraarrow] ([yshift=2mm]none.east) -- node[agralabel, above, align=center]
    {grant$(x,o)$ [teacher, $I{=}S{=}\T$]\\/ $v_\pi{:=}v$, $d{:=}t{+}\ell$}
    ([yshift=2mm]live.west);
  \draw[agraarrow] ([yshift=-2mm]live.west) -- node[agralabel, below, align=center]
    {revoke; consume [adm];\\tick [$t{>}d$]; update [adm] / $v{:=}v{+}1$}
    ([yshift=-2mm]none.east);
  \path[agraarrow] (none) edge[loop above, min distance=12mm] node[agralabel, above, align=center]
    {tick; update [adm] / $v{:=}v{+}1$} (none);
  \path[agraarrow] (live) edge[loop above, min distance=12mm] node[agralabel, above, align=center]
    {grant (replace); tick [$t{\le}d$]} (live);
\end{tikzpicture}
\caption{Permit and policy states, with transitions labeled
\emph{event [guard] / effect}; $I$ and $S$ are the grant's identity and source evidence.
\emph{Consume} and \emph{update} are successful commits.
Events not drawn change no component shown.}\label{fig:permit}
\end{figure}

Four events change the state.
\emph{Grant} creates a
permit $(\mathit{yes},x,o,v,t+\ell)$ for actor $x$ and
operation $o\in\{\code{TOOL\_CALL},\code{EXPORT\_DATA}\}$;
it requires the teacher role and confirmed identity and source evidence.
\emph{Revoke} clears the permit.
\emph{Tick} advances $t$ and deactivates the permit if past the deadline.
\emph{Commit}
re-evaluates a candidate in the current state and, if it is admissible,
applies its effect: an update installs the replacement policy and revokes
the permit, a protected operation consumes the permit,
and an observation has no external effects.
A rejected
commit leaves the state unchanged.
The implementation also counts executions and accepted updates;
these counters are read only by overflow guards and are omitted here.
Selection among candidates (Sec.~\ref{sec:03b-semantics}) is pure;
only commit, grant, revoke, and tick change the state.
Fig.~\ref{fig:permit}
shows the resulting permit life cycle.

\emph{Assumptions and threat model.}
\emph{A1 (trusted adapters):} roles, identity and source evidence,
and data labels are produced by adapters outside the model, never copied from
model-generated text, and each candidate's role is that of its actor.
\emph{A2 (evaluator integrity):} the attacker cannot modify the evaluator,
the hard properties, or the state directly.
\emph{A3 (atomic commit):} revalidation and effect of a commit are atomic.

The attacker can propose arbitrary candidates, supply external content,
replay stale expected versions, and exploit missing or conflicting evidence.
Under A1--A3, SFR guarantees the invariants of Sec.~\ref{sec:03d-guarantees}.
If an adapter errs, the affected property fails open: a falsely confirmed
identity defeats the identity check, and confidential data labeled public
defeats the flow check.
The other properties still apply, so a single
adapter error does not by itself enable a policy update by a non-teacher.

\subsection{Semantics}\label{sec:03b-semantics}

Admission ($\adm$) depends on how evidence is represented, which properties are checked,
and how admissible candidates are ranked.
We define these three layers in turn and explain where the four-valued
representation changes outcomes.

\emph{Four-valued evidence.}
An evidence value is a pair $(e^+,e^-)\in\{0,1\}^2$ of positive and negative
support (Table~\ref{tab:values}).
Negation swaps the components; conjunction and disjunction are
\begin{eqnarray}
 a\wedge b&=&(a^+\land b^+,\ a^-\lor b^-),\label{eq:and}\\
 a\vee b&=&(a^+\lor b^+,\ a^-\land b^-).
\end{eqnarray}
These operations combine \emph{requirements} in the truth order.
Observations of the \emph{same} proposition from several adapters are combined
by the knowledge join
\begin{equation}\label{eq:merge}
 a\oplus b=(a^+\lor b^+,\ a^-\lor b^-),
\end{equation}
so a confirming and a refuting observation yield $\B$.
Boolean checks enter the algebra through the embedding $\mathit{true}\mapsto\T$,
$\mathit{false}\mapsto\F$.

\begin{table}[t]
\caption{Four-valued evidence and admission (left) and the bilattice FOUR
with truth order $\le_t$ and knowledge order $\le_k$ (right).
}\label{tab:values}
\centering
\begin{tabular}[c]{c c c l c}
\hline
Value & $e^+$ & $e^-$ & Meaning & Admits\\\hline
$\N$ & 0 & 0 & Unknown & No\\
$\T$ & 1 & 0 & Confirmed & Yes\\
$\F$ & 0 & 1 & Refuted & No\\
$\B$ & 1 & 1 & Inconsistent & No\\\hline
\end{tabular}\hspace{2.5cm}
\begin{tikzpicture}[baseline=(c.base), x=1.5mm, y=1.5mm,
  v/.style={draw=none, fill=agralightblue, minimum width=7mm, minimum height=6mm,
    inner sep=0pt, font=\normalsize},
  agralabel/.append style={font=\small}]
  \node[v] (N) at (0,0) {$\N$};
  \node[v] (F) at (-6.5,6.5) {$\F$};
  \node[v, fill=agrablue, text=white] (T) at (6.5,6.5) {$\T$};
  \node[v] (B) at (0,13) {$\B$};
  \coordinate (c) at (0,5.5);
  \draw[agrablue, line width=0.5pt] (N) -- (F) -- (B) -- (T) -- (N);
  \draw[agraarrow] (-11,-2) -- node[agralabel, left] {$\le_k$} (-11,14);
  \draw[agraarrow] (-9,-4.5) -- node[agralabel, below] {$\le_t$} (9,-4.5);
\end{tikzpicture}
\end{table}

Let $P_1,\dots,P_6$ be the hard properties of Table~\ref{tab:properties}.
Then,
\begin{equation}\label{eq:admission}
 \adm(s,a)\iff P_i(s,a)=\T\ \text{for all } i\in\{1,\dots,6\}.
\end{equation}
Designating only $\T$ rejects refuted, unknown, and inconsistent evidence alike.
A gate that checks $e^+$ alone would designate $\{\T,\B\}$ and admit conflicts;
the implementation therefore compares complete values.

\emph{Role of four-valued evidence.}
For execution, Eq.~(\ref{eq:admission}) is a Boolean gate.
The four values matter in three other places.
\emph{Merging:} by Eq.~(\ref{eq:merge}), a single refuting observation turns $\T$
into $\B$ and withdraws permission, so authorization is not monotone in
accumulated evidence.

\emph{Feedback:} the property vector separates refutation,
which yields a counterexample, from missing evidence,
which calls for further observation, and from conflicting evidence,
which calls for conflict resolution. 
Feedback must use the vector, not the conjunction of its entries,
because $\N\wedge\B=\F$ by Eq.~(\ref{eq:and}): conjoining unknown and conflicting
evidence would report a refutation.

\emph{Composition:} properties combine evidence with structural checks by
$\wedge$, without ad-hoc rules.
A Boolean gate with reason codes could reproduce the feedback labels,
but it would need its own rules for merging and composition;
the lattice supplies them with known algebraic laws.

\emph{Hard properties.}
Table~\ref{tab:properties} lists the six properties.
An inactive property yields $\T$.
For updates, integrity is the conjunction of source evidence and a structural
check of the replacement: bounds, successor version $v+1$,
and an overflow guard.
With $C$ for confidential and $P$ for public, the flow property is
\begin{equation}\label{eq:flow}
 \mathit{flow}(L_{\rm in},L_{\rm sink})=
 \begin{cases}
 \F, & L_{\rm in}=C \text{ and } L_{\rm sink}=P,\\
 \T, & \text{otherwise}.
 \end{cases}
\end{equation}
A permit is \emph{live} for candidate $a$ if it is active,
its actor and kind match $a$, $v_\pi=v$, and $t\le d$.
A live permit does not override Eq.~(\ref{eq:flow}), and the model has no
declassification.

\begin{table}[b]
\caption{Hard properties (U: update, P: tool call or
export).}\label{tab:properties}
\centering\small
\begin{tabular}{>{\raggedright\arraybackslash}p{.24\columnwidth} c >{\raggedright\arraybackslash}p{.52\columnwidth}}
\hline
Property & Active & Requirement\\\hline
$P_1$ Identity & U, P & Identity evidence is $\T$.\\
$P_2$ Role & U, P & Teacher for U; teacher or student for P.\\
$P_3$ Integrity & U & Source evidence $\wedge$ structural check is $\T$.\\
$P_4$ Flow & P & Eq.~(\ref{eq:flow}) yields $\T$.\\
$P_5$ Permit & P & A live permit matches.\\
$P_6$ Freshness & all & Expected version equals $v$.\\\hline
\end{tabular}
\end{table}

\begin{figure}[t]
\centering
\begin{tikzpicture}[node distance=8mm and 14mm,
                    agrabox/.append style={font=\normalsize, minimum height=10mm}]
  \node[agrabox, text width=30mm] (cand) {Candidate\\proposal};
  \node[agrabox, text width=30mm, right=of cand] (eval) {Property\\evaluation};
  \node[agrabox, text width=30mm, right=of eval] (rank) {Admissible\\ranking};
  \node[agrabox, text width=30mm, below=of eval] (state) {Trusted state\\and evidence};
  \node[agrabox, text width=30mm, below=of rank] (commit) {Revalidate\\and commit};
  \node[agrabox, text width=30mm, below=of cand] (fb) {Feedback\\to raising};
  \draw[agraarrow] (cand) -- (eval);
  \draw[agraarrow] (eval) -- (rank);
  \draw[agraarrow] (rank) -- (commit);
  \draw[agraarrow] (commit) -- (state);
  \draw[agraarrow] (state) -- (eval);
  \draw[agraarrow] (eval.south west) -- (fb.north east);
\end{tikzpicture}
\caption{Evaluation pipeline. A proposal cannot change trusted state directly; only an admitted, revalidated commit does. Feedback is derived from the property vector and explicitly not from the model's self-assessment.}\label{fig:flow}
\end{figure}
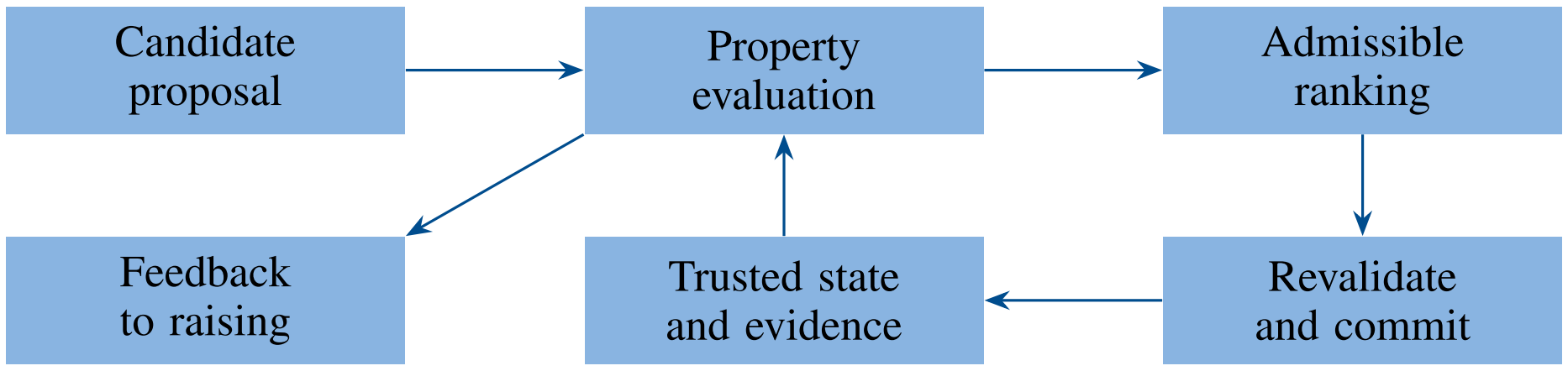

\emph{Preferences and selection.}
Admissible candidates are compared by two binary costs,
for privacy and for learning progress.
Priorities are ordinal: the candidate with the lower total cost at the highest
priority level
where the candidates differ is preferred, and costs at equal priority are
summed.
Selection computes
the admissible set and returns its unique minimum.
Several minima yield \code{UNDERDETERMINED}
and an empty set yields \code{NO\_ADMISSIBLE}; neither selects a candidate.
A selection result
is not a permission: commit re-evaluates Eq.~(\ref{eq:admission}) in the state
current at that time.
Fig.~\ref{fig:flow} summarizes the pipeline.

\subsection{Illustrative Scenarios}\label{sec:03c-scenarios}
To explain the model's behavior, we show scenarios, using
the same evaluator instead of scenario-specific rules. 

\emph{An external policy injection.}
An external source proposes a policy replacement.
Its identity may be known and the replacement may be structurally valid,
but the external role is not authorized to change the policy and source trust is
refuted.
The candidate is blocked with role and integrity failures.
A second candidate from a confirmed teacher with trusted provenance and the next
valid version is admitted.
Selection chooses the teacher update.
Persuasive content or a low preference cost cannot repair the external
candidate's hard failures.

\emph{Conflicting identity and unknown provenance.}
A proposed update has the teacher role but identity evidence $\B$.
The identity requirement blocks it, and the feedback requests teacher review.
In another scenario, identity is confirmed but source trust is $\N$.
Integrity remains unknown, and the update is again blocked with teacher-review
feedback.
These examples show why keeping the evidence values is useful even when both
cases lead to the same execution decision.

\emph{A valid permit does not permit every flow.}
A student holds an active export permit.
The proposed export carries confidential data to a public destination.
Live authorization is satisfied, but the flow property is refuted.
The transfer is blocked.
This illustrates the conjunction of independent requirements:
permission to invoke an operation does not remove its data-handling constraints.

\emph{Preferences among secure alternatives.}
Two public-data exports satisfy all hard requirements and share a matching
permit.
Candidate $a$ has privacy cost 1 and progress cost 0; $b$ has the reverse.
At the default priorities, $b$ is preferred because privacy has the higher
priority.
If progress received the higher priority, $a$ would be preferred.
This is a choice among admissible abstract behaviors.
The second candidate is not executed merely because it was evaluated,
and a successful commit would consume the shared permit.

\emph{Revocation and underdetermination.}
After an export is selected, the teacher revokes its permit.
Commit re-evaluates authorization and blocks execution.
This exposes the distinction between a decision made under old state and a
transition attempted under current state.
Finally, two inert observations with identical preference costs yield
\code{UNDERDETERMINED}.
The evaluator does not invent a distinction absent from the specified policy.

\subsection{Guarantees}\label{sec:03d-guarantees}
We state the security invariants of SFR and prove them by induction,
show that only $\T$ may admit an action,
and identify a gap that the invariants do not cover.
Section~\ref{sec:04-implementation} checks the invariants on the reference
implementation.

A state is \emph{reachable} if it results from the initial state by a finite
sequence of grant, revoke, tick, and commit events.

\begin{theorem}[Security invariants]\label{thm:inv}
Under A1--A3, every step from a reachable state satisfies:
\begin{itemize}[leftmargin=*, labelindent=0pt, widest=I6]
\item[I1] \emph{Authorization:} a successful protected commit by actor $x$ for
operation $o$ occurs only if the permit is live for $x$ and $o$.
\item[I2] \emph{Single use:} after a successful protected commit,
no permit is active.
\item[I3] \emph{Explicit flow:} no successful protected commit moves
confidential data to a public sink.
\item[I4] \emph{Policy integrity:} $v$ changes only by a successful update from
a teacher with identity and source evidence $\T$; the new version is $v+1$ and
the installed policy is within bounds.
\item[I5] \emph{Version binding:} every active permit satisfies $v_\pi=v$.
\item[I6] \emph{Conservative gate:} no successful commit has a property value in
$\{\N,\B\}$, and rejected commits and grants leave the state unchanged.
\end{itemize}
\end{theorem}

\begin{IEEEproof}[Proof]
I1, I2, I3, I6, and the first part of I4 follow from Eq.~(\ref{eq:admission}),
Table~\ref{tab:properties}, and the effects of commit,
because commit re-evaluates every property
in the current state.
For I5 and the bounds in I4 we use induction over steps.
The initial
policy is valid and no permit is active.
A grant binds $v_\pi=v$.
An accepted update
increments $v$, passes the structural check, and revokes the permit.
No other event changes $v$ or the policy.
\end{IEEEproof}

Admission designates $\T$ only.
The following lemma shows that this choice is forced by the invariants.

\begin{lemma}[Only confirmed evidence admits]\label{lem:designated}
Let $D$ be a set of evidence values with $\T\in D$,
and let $\adm_D(s,a)$ hold iff $P_i(s,a)\in D$ for all $i$.
Then I1--I6 hold under $\adm_D$ if and only if $D=\{\T\}$.
\end{lemma}

\begin{IEEEproof}
For $D=\{\T\}$, $\adm_D$ is Eq.~(\ref{eq:admission}) and Theorem~\ref{thm:inv}
applies.
Otherwise $D$ contains some $x\neq\T$.
In the initial state, consider a policy update by the teacher role
with identity evidence $x$, source evidence $\T$, the current expected version,
and a valid successor policy.
Then $P_1=x$ and $P_2,\dots,P_6$ are $\T$, so the update is admitted and commits
although its identity evidence is not $\T$, which violates I4.
\end{IEEEproof}

Neither the paraconsistent choice $D=\{\T,\B\}$
nor the gap-tolerant choice $D=\{\T,\N\}$ is compatible with the invariants.

\emph{A confused-deputy gap and its repair.}
The invariants do not constrain the source of a protected operation:
$P_3$ is inactive for tool calls and exports.
Imagine a scenario where a student holds a tool permit
and an injected web page causes it to propose a tool call;
the candidate carries the student's identity and role and the live permit,
so it is admitted although its source is untrusted.
This is the confused deputy of~\cite{hardy}.
Base SFR therefore protects the \emph{policy} against injected content,
but not \emph{tool use}. The repair activates $P_3$ for protected operations 
and requires source evidence $\T$ there as well.
With this extension, a protected operation requires confirmed provenance.
The price is that adapters must report the provenance of the instruction
that caused a call, which is what capability-based data-flow tracking
provides~\cite{camel}.

\section{Reference Implementation and Conformance}\label{sec:04-implementation}
The C artifact requires neither a solver nor a language-model API.
The structures \code{Policy}, \code{State}, \code{Action},
and \code{Evaluation} correspond to the components above.
The function \code{evaluate} returns six evidence values and an admission flag;
\code{select\_action} performs filtering and ranking;
\code{commit} revalidates and applies a transition.
Grant, revoke, and clock-advance events have separate functions.

Figure~\ref{fig:code} shows the compact evidence representation and the
essential admission test.
The negative support mask uses bit 1, so the integer 2 represents refutation;
it is not a severity weight.
The feedback is derived from the property outcomes,
not from the model's self-assessment.
\begin{figure}[t]
\begin{lstlisting}[style=agra, escapeinside={(*@}{@*)}]
typedef enum { N=0, T=1, F=2, B=3 } MV;
/* bit 0: e+ (support for T), bit 1: e- (for F) */
static MV mv_and(MV a, MV b) {
  return (MV)((a & b & 1) | ((a | b) & 2));
}
/* P3 for an update: source evidence AND
   structural check, embedded as T/F */
e.values[INTEGRITY] = mv_and(a->trusted_source,
    valid_policy(&a->replacement)
    && s->policy.version < UINT_MAX
    && a->replacement.version
       == s->policy.version + 1 ? T : F);
/* Eq. ((*@\ref{eq:admission}@*)): all COUNT = 6 properties T */
e.admissible = 1;
for (i = 0; i < COUNT; ++i)
  if (e.values[i] != T) e.admissible = 0;
\end{lstlisting}
\caption{Excerpt from the C implementation: evidence encoding, conjunction as in Eq.~(\ref{eq:and}),
composition of source evidence with a Boolean structural check (embedded as $\T$/$\F$),
and admission as in Eq.~(\ref{eq:admission}).}\label{fig:code}
\end{figure}

The feedback routine returns \code{positive-example} for an admissible
candidate.
If admission fails and at least one hard property is exactly $\F$,
it returns \code{counterexample}.
Otherwise, it returns \code{teacher-review}.
Thus, a definite failure takes precedence over unresolved evidence in the coarse
feedback label.
The full property vector remains available for diagnosis;
an inconsistent value $\B$ is not silently converted to either truth or
refutation.

With six fixed properties, evaluation costs constant time per candidate and
selection is linear in the number of candidates.
Storage for state and each candidate is fixed.
This complexity concerns the reference reasoner only:
obtaining identity or semantic evidence may be substantially more expensive.

A rejected grant preserves the permit; a valid grant replaces it.
Rejected commits preserve all state fields.
Invalid enumeration values and costs are rejected,
and successor-version and deadline arithmetic are guarded.
Candidates are constructed internally, without an untrusted text parser.

\subsection{Exhaustive Bounded Admission Enumeration}
The test mode constructs the Cartesian product of identity evidence (4 values),
source evidence (4), role (3), operation (4), input label (2),
destination label (2), and permit availability (2).
The total is $4\cdot4\cdot3\cdot4\cdot2\cdot2\cdot2=1{,}536$.
Every case starts from the default valid policy and uses the current expected
version and a structurally valid replacement.
For tool/export cases with availability enabled, the test creates a matching
fresh permit through the trusted grant function.
For observations and updates, the availability flag duplicates otherwise
identical cases because permits are irrelevant there.

A separately expressed Boolean oracle is compared with evaluation and commit
results.
Rejected commits are checked for preservation of every state field.
The oracle does not use evaluator helpers, but both can share specification
errors.

Table~\ref{tab:enumeration} partitions the enumerated domain.
Of 1,536 cases, 440 are admissible and 1,096 are blocked.
Among the blocked cases, 160 have teacher-review feedback and 936 have
counterexample feedback.
These counts characterize the constructed domain; they are not probabilities of
successful or malicious activity.
\begin{table}[t]
\caption{Admission outcomes in the bounded enumeration.}\label{tab:enumeration}
\centering
\begin{tabular}{l r r r}
\hline
Operation & Cases & Admitted & Blocked\\\hline
Observation & 384 & 384 & 0\\
Policy update & 384 & 8 & 376\\
Tool call & 384 & 24 & 360\\
Data export & 384 & 24 & 360\\\hline
Total & 1,536 & 440 & 1,096\\\hline
\end{tabular}
\end{table}

Observations are intentionally inert.
Updates require confirmed identity and provenance plus teacher role.
Tools and exports require confirmed identity, an internal role, a permit,
and one of three allowed label pairs; their provenance dimension is inactive.
These activation rules explain the counts.

\subsection{Algebra, Sequences, and Boundary Cases}
Further assertions cover double negation, commutativity of conjunction,
De Morgan's law, a material-implication case, and mixed evidence.
Sequence checks exercise single-use permits, revocation before commit,
the inclusive deadline, expiry after the deadline, actor and operation binding,
stale versions, invalidation after an update, and rejected rollback.
Preference tests cover ties, priority reversal, dominance of hard constraints,
and empty candidate sets.
Boundary tests cover clock and deadline overflow and invalid evidence values.

The run reports 4,617 passed assertions, including setup and state-preservation
checks, rather than 4,617 distinct properties.

\subsection{Bounded Exhaustive Exploration}
To check the theorem against the reference implementation,
a harness compiles the unchanged C code and drives its own functions.
A breadth-first search enumerates all states reachable from the initial state.
In each state it applies every tick and revoke, every grant
(all issuer roles, evidence values, actors, and operations),
and commits every candidate.
Candidates range over all operations, roles, three actor identifiers,
evidence values, and labels, over current and stale expected versions,
and over valid and invalid replacement policies;
roles and actors are combined freely, including combinations that A1 excludes.
Bounds are $t\le3$, $v\le4$, $\ell\in\{1,2\}$, and priorities in $\{30,40\}$;
states that differ only in the execution and update counters,
which only overflow guards read, are identified.
The search visits 6,262 reachable states and checks 72,329,329 transitions;
none violates I1--I6.
To guard against vacuous checks, we seeded faults into copies of the code.
Admitting $\B$ is detected by I4 and I6,
and keeping the permit after an update by I5.
Removing the deadline check from the permit test alone is undetectable,
because a tick already revokes an expired permit;
removing both deadline checks is detected by I1.
The search also counts successful protected commits whose source evidence is not
$\T$,
the confused deputy of Sec.~\ref{sec:03d-guarantees}:
it finds 21,708 such transitions.
With the extension it finds none, and I1--I6 still hold.
The extension is implemented as a second version of the reference implementation
that adds a single line to \code{evaluate};
in its enumeration, 6 instead of 24 tool calls and 6 instead of 24 exports are
admitted.
The three kinds of checks differ in scope:
the exploration covers all reachable states within its bounds.
Enumeration exhausts one base state's input product;
sequence checks sample histories.
None of these checks is a C correctness proof.
No deployed identity service, language model, or network tool is evaluated.

\section{Discussion and Direction for Future Work}\label{sec:05-discussion}
\emph{What SFR covers.}
Under A1--A3, Theorem~\ref{thm:inv} covers the policy-update channel,
the authorization of tool calls and exports,
and the freshness of proposals and permits.
With the extension of Sec.~\ref{sec:03d-guarantees},
it also covers protected operations triggered by untrusted content.
All guarantees are conditional on A1:
SFR specifies what adapters must deliver but does not deliver it.
The model deliberately keeps the protected state small.
The aspects below are possible extensions of the approach:
each adds operations, properties, or state components,
while the general concept remains unchanged.

\emph{Extensions of the model.}
The protected policy currently consists of the preference priorities
and the permit lifetime; FR's property set and hard flags are fixed.
Protecting their updates adds structural checks to $P_3$
but no new mechanism.
Review answers and training material flow back into raising outside the model;
treating curriculum changes as a further protected operation,
with role and source requirements as for policy updates,
covers this channel.
One permit record and an atomic, single-threaded commit keep the state small;
several permits and concurrent tools require a set of permits
and defined semantics for consumption and partial failure.
The two-label flow check generalizes to a lattice of labels~\cite{denning},
and a teacher-authorized declassification can be added
as a permit-like operation.
Finally, an authenticated teacher is trusted;
requiring several teachers to approve an update
adds one property of the same kind.

\emph{Who may contribute evidence.}
Because observations are merged by Eq.~(\ref{eq:merge}),
any accepted refuting observation turns $\T$ into $\B$ and blocks the action.
This is the intended conservative behavior, but it gives a veto to every party
whose observations are merged.
Adapters must therefore restrict, per proposition, who may contribute evidence,
and teacher review must resolve $\B$ within a bounded time;
otherwise a single source can deny service.
The same applies to repeated conflicts in general.

\emph{Relation to raising.}
SFR provides the environment-side guarantees of raising and a
feedback signal: admitted actions are positive examples,
refuted properties are counterexamples,
and missing or conflicting evidence is routed to a teacher.
Whether training on these traces
changes a learner's proposals has not been tested.
A first experiment would train a small
model on scenario traces and measure how often it still proposes actions that
SFR blocks.

\section{Conclusion}\label{sec:06-conclusion}
\emph{Secure Formal Raising} extends the property-oriented perspective of FR to the
environment in which an AI system develops and acts.
The reference model protects preference-policy updates,
scopes and revokes tool authorization, and blocks confidential-to-public
transfers.
Four-valued evidence retains uncertainty and conflict while admitting only
confirmed hard requirements.
Preferences rank admissible alternatives without creating exceptions to security
constraints.

We proved six security invariants and showed that they force admission
to accept only confirmed evidence:
designating unknown or inconsistent evidence as well admits a policy change
without confirmed identity.
A plain-C reference implementation makes the model executable;
a bounded exhaustive exploration confirms the invariants for it
and exposes a confused-deputy gap for tool use,
which a provenance requirement closes.
The results concern an abstract environment and do not establish security of
arbitrary language-model internals or a deployed raising pipeline.
Future work will connect the interfaces to real adapters
 and study how property traces support
curriculum construction and teacher feedback.

\bibliographystyle{unsrt}
\bibliography{literature}

\end{document}